\documentclass{llncs}
\usepackage{makeidx}  % allows for indexgeneration
\usepackage{graphicx}

\usepackage{siunitx}
\usepackage{amsmath,amsfonts,amssymb}
\usepackage{caption}
\usepackage{subcaption}
\usepackage{color}
\usepackage{arydshln}

\begin{document}
\sloppy
\pagestyle{headings}  % switches on printing of running heads
\title{Hydranet: Data Augmentation \\ for Regression Neural Networks}

\author{Florian Dubost\inst{1} \and Gerda Bortsova\inst{1} \and Hieab Adams\inst{1,2} \and M. Arfan Ikram\inst{1,2,3} \and Wiro Niessen\inst{1,4} \and Meike Vernooij\inst{1,2} \and Marleen de Bruijne\inst{1,5}}

\institute{Department of Radiology \& Nuclear Medicine, Erasmus MC - University Medical Center Rotterdam, the Netherlands \and
Department of Epidemiology, Erasmus MC, Rotterdam, the Netherlands \and
Department of Neurology, Erasmus MC, Rotterdam, the Netherlands \and
Department of Imaging Physics, Faculty of Applied Science, TU Delft, Netherlands \and
Department of Computer Science, University of Copenhagen, Denmark}

\maketitle

\begin{abstract}
Deep learning techniques are often criticized to heavily depend on a large quantity of labeled data. This problem is even more challenging in medical image analysis where the annotator expertise is often scarce. We propose a novel data-augmentation method to regularize neural network regressors that learn from a single global label per image. The principle of the method is to create new samples by recombining existing ones. We demonstrate the performance of our algorithm on two tasks: estimation of the number of enlarged perivascular spaces in the basal ganglia, and estimation of white matter hyperintensities volume. We show that the proposed method improves the performance over more basic data augmentation. The proposed method reached an intraclass correlation coefficient between ground truth and network predictions of 0.73 on the first task and 0.84 on the second task, only using between 25 and 30 scans with a single global label per scan for training. With the same number of training scans, more conventional data augmentation methods could only reach intraclass correlation coefficients of 0.68 on the first task, and 0.79 on the second task.
\end{abstract}

\section{Introduction}
Deep learning techniques are getting increasingly popular for image analysis but are often dependent on a large quantity of labeled data. In case of medical images, this problem is even stronger as data acquisition is administratively and technically more complex, as data sharing is more restricted, and as the annotator expertise is scarce. 

To address biomarker (e.g. number or volume of lesions) quantification, many methods propose to optimize first a segmentation problem and then derive the target quantity with simpler methods. These approaches require expensive voxel-wise annotations. In this work, we circumvent the segmentation problem by optimizing our method to directly regress the target quantity \cite{Cole2017,Dubost2018,Gonzalez2018,Wang2019,Lee2018}. Therefore we need only a single label per image instead of voxel-wise annotations. Our main contribution is that we push this limit even further by proposing a data augmentation method to reduce the number of training images required to optimize the regressors. The proposed method is designed for global image-level labels that represent a countable quantity. Its principle is to combine \textit{real} training samples to construct many more \textit{virtual} training samples. During training, our model takes as input random sets of images and is optimized to predict a single label for each of these sets that denotes the sum of the labels of all images of the set. This is motivated by the idea that adding a large quantity of virtual samples with weaker labels may reduce the over-fitting to training samples and improve the generalization to unseen data.

\subsection{Related Work}

Data augmentation can act as a regularizer and improve the generalization performance of neural networks. In addition to simple data-augmentations such as rotation, translation and flipping, the authors of Unet \cite{Ronneberger2015} stress for instance that random elastic deformations significantly improved the performance of their model.
Generative adversarial networks have for instance also been used to generate training samples, and hence reduce the over-fitting \cite{Sixt2018}.

Recently, data augmentation methods using combinations of training samples have been published.
Zhang et al. \cite{Zhang2018} proposed to construct virtual training samples by computing a linear combination of pairs of real training samples. The corresponding one-hot labels are summed with the same coefficients. The authors evaluated their method on classification datasets from computer vision and on a speech dataset, and demonstrate that their method improves the generalization of state-of-the-art neural networks. Simultaneously, Inoue et al. \cite{Inoue2018} and Tokozume et al. \cite{Tokozume2017} reached similar conclusions.
In case of grayscale volumetric inputs, summing image intensity values could overlay the target structures, confuse discriminative shapes, and thus harm the performance of the network. With our method, training samples can be combined without overlaying the intensity values.
The other difference with the above-mentioned approaches is that our method is also not designed for classification, but for regression of global labels, such as volume or count in an image. With the proposed combination of samples, our method computes plausible augmentation.

\section{Methods}
\label{sec:methods}

\begin{figure}[t]
\centering
\includegraphics[height=4cm]{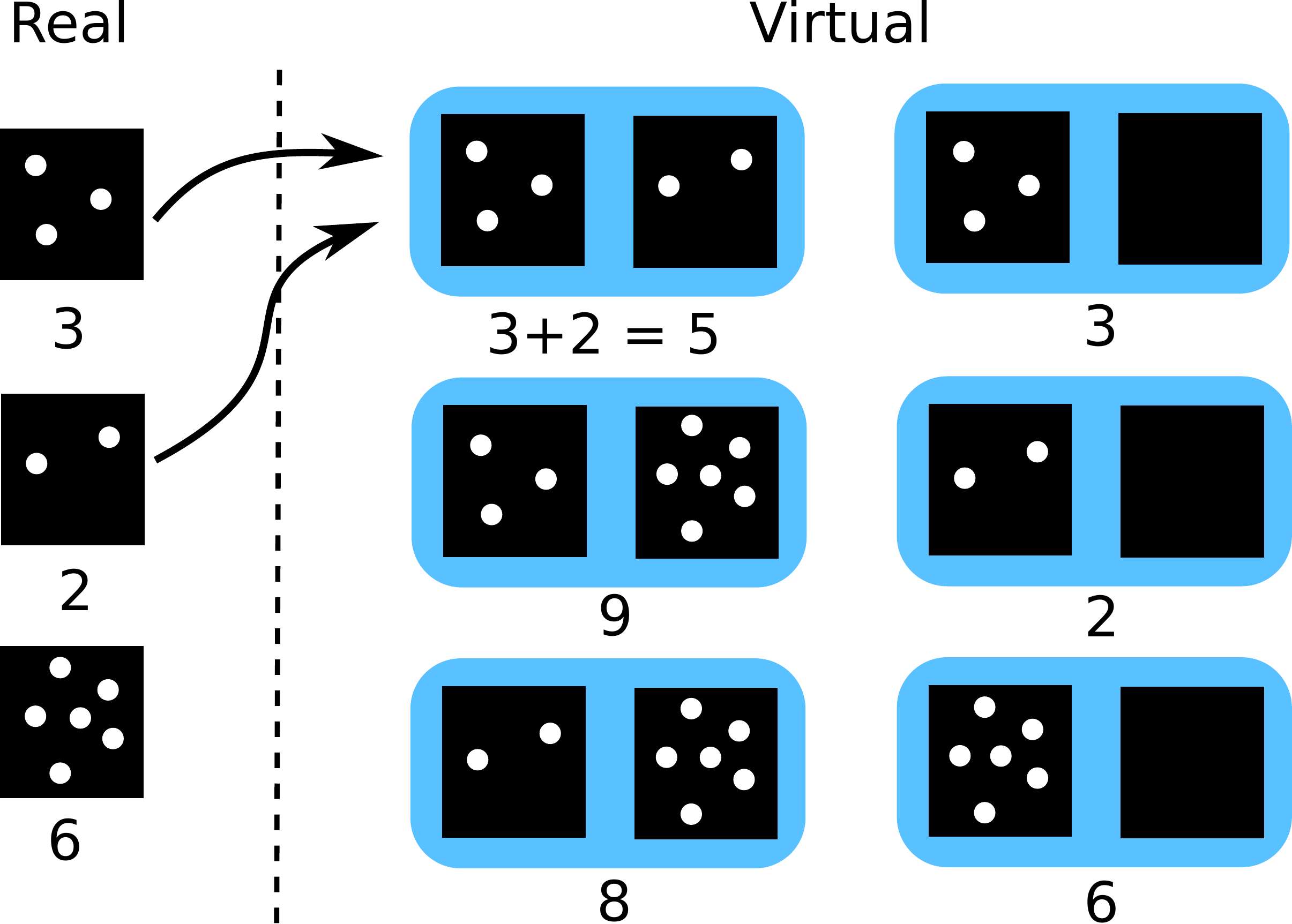}
\caption{\textbf{Creating virtual training samples by recombining real training samples for regression tasks.} The real training samples are displayed on the left, and the virtual samples on the right. The label is indicated under each sample, and corresponds to the number of white blobs. By recombining samples, we can significantly increase the size of the training dataset. For example, by recombining the real samples with labels 3 and 2, we can create a new sample with label 5 (arrows). All possible combinations are shown in blue. For the illustration, we show only combinations of two samples, but any number of samples can be combined. In our experiments, we used combinations of maximum 4 samples.}
\label{fig:principle}
\end{figure}

The principle of the proposed data augmentation method is to create many new (and weaker) training samples by combining existing ones (see Figure \ref{fig:principle}). In the remainder, the original samples are called \textit{real samples}, and the newly created samples are called \textit{virtual samples}.

\subsection{Proposed Data Augmentation.}
\label{sec:principle}

During training, the model is not optimized on single real samples $I$ with label $y$, but on sets $S$ of $n$ random samples $I_{1},I_{2},...,I_{n}$  with label $y_{s} = \sum\limits_{i=1}^n y_{i}$, with $y_{i}$ the label of sample $I_{i}$. These sets $S$ with labels $y_{s}$ are the virtual samples. Consequently, the loss function $L$ is computed directly on these virtual samples $S$ and not anymore the individual real samples $I_{i}$. This approach is designed for labels describing a quantitative element in the samples, such as volume or count in an image.

To create the sets $S$, the samples $I_{i}$ are drawn without replacement from the training set at each epoch. To create more combinations of samples, and to allow the model to use the real samples for its optimization, the size of the sets $S$ can randomly vary in $\{1,n\}$ during training. If the training set contains $m$ samples, with our method, we can create $\sum\limits_{i=1}^{n} \binom{m}{i}$ possible different combinations (the order of the samples $I_{i}$ in $S$ has no effect on the optimization).

\paragraph{Difference with mini-batch stochastic gradient descent (SGD).}
In mini-batch SGD, the model is also optimized on sets of random samples, but the loss function $L$ is computed individually for each sample of the batch, and then summed (averaged). For the proposed method, the predictions are first summed, and the loss function is then computed a single time. For non-linear loss functions, this is not equivalent: $\sum\limits_{i=1}^{n} L(\hat{y}_{i},y_{i}) \neq L(\sum\limits_{i=1}^{n} \hat{y}_{i}, \sum\limits_{i=1}^{n} y_{i})$, with $\hat{y}_{i}$ the model's prediction for sample $I_{i}$.

\paragraph{Regularization Strength}
The regularization strength can usually be modulated by at least one parameter, for instance the degree of rotation applied to the input image, or the percentage of neurons dropped in Dropout \cite{Srivastava2014}. In the proposed method, the regularization effect can be controlled by varying the average number of samples used to create combinations.

\subsection{Implementation.}
\label{sec:implementation}

We optimize a regression neural network with a 3D image for input, and global label representing a volume or count for output.
There are at least two possible implementations of the proposed method. The first implementation could consist of modifying the computation of the loss function across samples in a mini-batch, and provide mini-batches of random size. Alternatively the model’s architecture could be adapted to receive the set of images. We opted for the second approach. 
 
\begin{figure}[t]
\centering
\includegraphics[height=5.2cm]{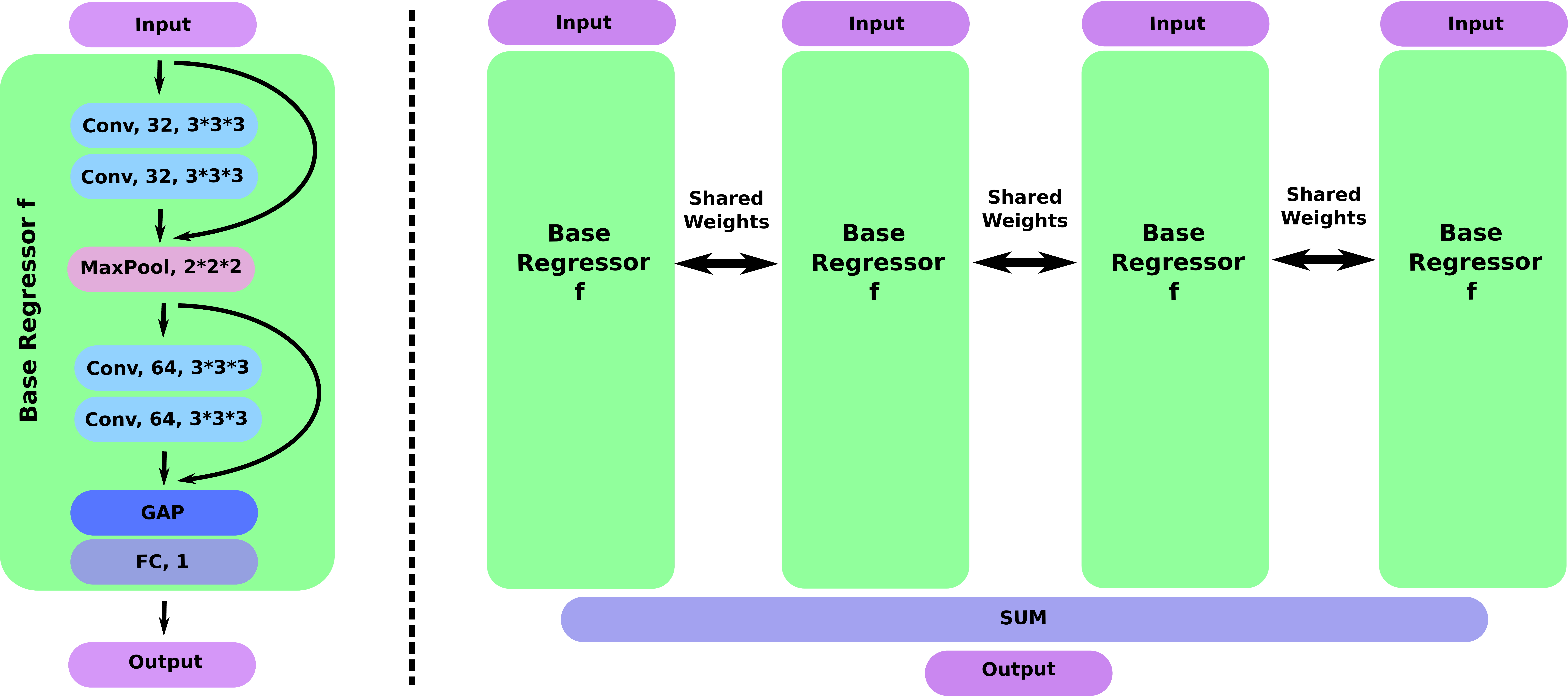}
\caption{\textbf{Architectures}. On the left, architecture of the base regressor $f$. 'Conv' stands for 3D convolutions, followed by the number of feature maps, and the kernel size. After each convolution, there is a ReLU activation. The round arrows are skip connections with concatenated feature maps. GAP stands for Global Average Pooling layer, and FC for Fully Connected layer. On the right, example of our data augmentation method with $n = 4$ replications. Each replication $f_{i}$ is a copy of the base regressor $f$ on the left. Once the training is done, all $f_{i}$ but one can be removed, and the evaluation is performed using the original architecture.}
\label{fig:arch}
\end{figure}

\paragraph{Base Regressor.} Figure \ref{fig:arch} left shows the architecture of the base regression neural network. It is both simple (196 418 parameters) and flexible to allow fast prototyping. There is no activation function after the last layer. The output $\hat{y}$ can therefore span $\mathbb{R}$ and the network is optimized with the mean squared error (MSE). We call this regression network $f$, such that $f(x)=\hat{y}$, with $x$ the input image.

\paragraph{Combination of Samples.}To process several images simultaneously, we replicate $n$ times the regressor $f$ during training (Figure \ref{fig:arch} right), resulting in $n$ different branches $f_{1}, f_{2},..., f_{n}$ that receive the images  $I_{1},I_{2},...,I_{n}$. The weights of each head $f_{i}$ are shared such that $f_{i} = f$. A new network $g$ is constructed as:

\begin{equation}
g(S) = g(I_{1},I_{2},...,I_{n}) = \sum\limits_{i=1}^n f_{i}(I_i) = \sum\limits_{i=1}^n f(I_i) = \sum\limits_{i=1}^n \hat{y}_{i}.
\end{equation}

To allow the size of the sets $S$ to randomly vary in $\{1,n\}$ during training, each element of $S$ has a chance $p$ to be a black image $B$ of zero intensities only (Figure \ref{fig:principle} right column). With $f(B) = 0$, the following situation becomes possible:
\begin{equation}
g(S) = f(I_{j}) + \sum\limits_{i=1, i \neq j}^n f_{i}(B) = f(I_{j}) + (n-1)f(B) = f(I_{j}).
\end{equation}

For this implementation, the batch size $b$ has to be a multiple of the number of branches $n$. We chose $b=n$ due to constraints in GPU memory. 
The regularization strength is controlled by the averaged number of samples used to create combinations, hence depends on $n$ and $p$. During inference, to predict the label for a single input image, the input of all other branches is set to zero.

\section{Experiments}
\label{sec:exp}
Enlarged perivascular spaces (PVS) and white matter hyperintensities (WMH) are two types of brain lesions associated with small vessel disease.
The method is evaluated for the estimation of number PVS in the basal ganglia, and estimation of WMH volume. We compare the performance of our method to that of the base regressor $f$ with and without and Dropout, and for different sizes of training set.

The PVS dataset contains T2-weighted scans, from 2017 subjects, acquired from a 1.5T GE scanner. The scans were visually scored by an expert rater who counted the PVS in the basal ganglia in a single slice.
The WMH dataset is the training set of the MICCAI2017's WMH challenge \cite{Kuijf2019}. We use the available 2D multi-slice FLAIR-weighted MRI scans as input to the networks. Scans were acquired from 60 participants from 3 centers: 20 scans from Amsterdam (GE scanner), 20 from Utrecht (Philips) and 20 from Singapore (Siemens). Although the ground truths of the challenge are pixel-wise, we only used the number of WMH voxels as ground truth during training.

For the regression of PVS in the basal ganglia, a mask of the basal ganglia is created with the subcortical segmentation algorithm from FreeSurfer \cite{desikan}, and smoothed with a gaussian filter (standard deviation of 2 voxels) before being applied the image. The result is subsequently cropped around the basal ganglia.
For the WMH dataset, we only crop each image around its center of mass, weighted by the voxel intensities.
For both tasks the intensities are then rescaled between 0 and 1.

During training, for all methods, the images are randomly augmented on-the-fly with standard methods. The possible augmentations are flipping in $x,y$ or $z$, 3D rotation from -0.2 to 0.2 radians and random translations in $x,y$ or $z$ from -2 to 2 voxels. Adadelta \cite{Zeiler2012} is used as optimizer.
The networks are trained with batch-size $b=4$. For the proposed method, the network's architecture has then four branches ($n=b=4$).
During an epoch, the proposed method gets as input $m/n$ different combinations of $n$ training samples, were $m$ is the total number of training images. During the same epoch, the base regressor $f$ simply gets the $m$ images separately (in batches of size $b=4$).
For the proposed method $p$ was set to 0.1.
In some experiments with Dropout  \cite{Srivastava2014} we included a dropout layer after each convolution and after the global pooling layer.
The code is written in Keras with Tensorflow as backend, and the experiments were run on a Nvidia GeForce GTX 1070 GPU.

For the PVS dataset, we experiment with varying size of training set, between 12 and 25 scans. The validation set always contains the same 5 scans. All methods are evaluated on the same separated test set of 1977 scans.
For the WMH dataset, the set is split into 30 training scans and 30 testing scans. Six scan from the training set are used as validation scans.
In both cases, the dataset is randomly (uniform distribution) split into training and testing sets. For the PVS dataset, once the dataset has been split into 30 training scans and 1977 testing scan, we manually sample scans to keep a pseudo-uniform distribution of the lesion count when decreasing the number of training scans.

To compare the automated predictions to visual scoring (for PVS) or volumes (for WMH), we use two evaluation metrics: the mean squared error (MSE), and the intraclass correlation coefficient (ICC).

\begin{figure}[t]
\centering
\includegraphics[height=3.3cm]{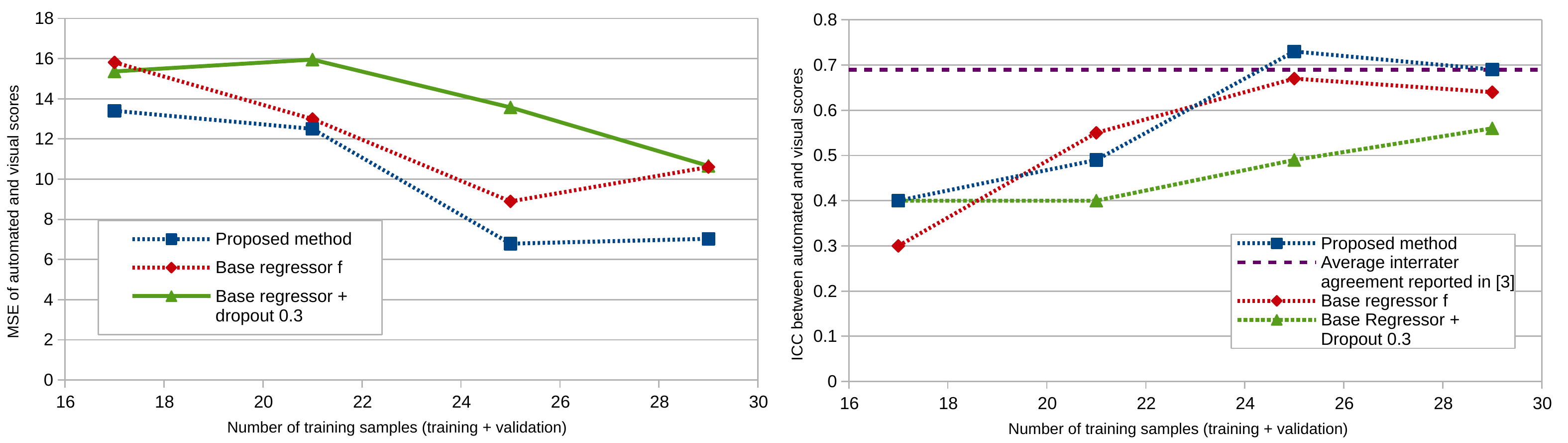}
\caption{\textbf{Comparison between the proposed method with $n=4$ and the base regressor on the PVS dataset.} MSE is displayed on the left, and ICC on the right.}
\label{fig:learningCurve}
\end{figure} %

\begin{table*}[h]
\setlength{\tabcolsep}{5pt}
\caption{\textbf{Results on the WMH dataset.} We conducted three series of experiments with different training set sizes and loss functions. In the two first rows, we repeated the experiments with three random initializations of the weights (on the same split), and report mean and standard deviation. MAE is an acronym for mean absolute error.}
\begin{center}
\begin{tabular}{c c c c c c}
\hline\rule{0pt}{12pt}
Method & Training scans & Testing scans & Loss & Performance (ICC) \\
\hline\rule{0pt}{12pt}
Base Network $f$    &    30    &    30    &    MSE &  0.79 $\pm$ 0.12 &\\
Proposed Method    &    30    &    30    &   MSE & \textbf{0.84 $\pm$ 0.02} &\\
\hdashline[0.5pt/3pt]\rule{0pt}{12pt}
Base Network $f$    &    30    &    30    &    MAE &  0.78 &\\
Proposed Method    &    30    &    30    &   MAE & \textbf{0.87} &\\
\hdashline[0.5pt/3pt]\rule{0pt}{12pt}
Base Network $f$    &    40    &    20    &    MSE &  \textbf{0.89} &\\
Proposed Method    &    40    &    20    &   MSE & 0.86 &\\
\hline\rule{0pt}{12pt}
\end{tabular}
\end{center}
\label{table:resultsWMH}
\end{table*}

\subsection{Results}
\label{sec:results}
\paragraph{Enlarged Perivascular Spaces (PVS).}
Figure \ref{fig:learningCurve} compares the proposed method to the base regressor $f$ on the PVS datasets, and for an increasing number of training samples. Their performance is also compared to the average interrater agreement computed for the same problem and reported in \cite{Dubost2018}.
The proposed method always reaches a better MSE than the conventional methods for all training set sizes. 
The proposed method also significantly outperforms the base regressor in ICC (Williams' test p-value $<$ 0.001) when averaging the predictions of the methods across the four points of their learning curve.

\paragraph{White Matter Hyperintensities (WMH).}
We conducted three series of experiments, and trained in total five neural networks (Table \ref{table:resultsWMH}). When using small training sets, the proposed method outperforms the base network $f$, when optimized either for MSE or for mean absolute error. With larger training sets, the difference of performance reduces, and the base regressor performs slightly better on the ICC.

\section{Discussion and Conclusion}

With the proposed data augmentation method, we could reach the inter-rater agreement performance on PVS quantification reported by Dubost et al. \cite{Dubost2018} with only 25 training scans, and without pretraining.

Dubost et al \cite{Dubost2018} also regressed the number of PVS in the basal ganglia with a neural network. We achieve a similar result (0.73 ICC) while training on 25 scans instead of 1000. Zhang et al. \cite{Zhang2018} also proposed to combine training samples as a data augmentation method. In their experiments, combining more than $n=2$ images does not bring any improvement. With the proposed method, training with combinations of four images brought improvement over only using pairs of images. We did not experiment with values of $n$ larger than 4 due to GPU memory constraints. Contrary to the expected gain in generalization, on both PVS (Figure \ref{fig:learningCurve}) and WMH datasets, using Dropout \cite{Srivastava2014} worsened the results when training on very little data, even with low dropout rates such as 0.3. As dropout already did not improve the performance of the baseline, we do not expect improvement by including dropout in the proposed method.

To create combination of images for the proposed method, images where drawn without replacement for the sake of implementation simplicity. The regularization strength could be increased by drawing samples with replacement, which could be beneficial for small training sets. We also mentioned two possible implementations of the proposed method: (1) changing the computation of the loss over mini-batches, (2) replicating the architecture of network. In this work we used the second approach, as it was simpler to implement with our library (Keras). However with this approach, all samples used in a given the combination have to be simultaneously processed by the network, which can cause GPU memory overload in case of large 3D images or large values of $n$. The first approach does not suffer from this overload, as the samples can be successively loaded, while only saving the individual scalar predictions in the GPU memory. In case of large 3D images, we would consequently recommend implementing the first approach.

\section{Acknowledgments}
This research was funded by the Netherlands Organisation for Health Research and Development (ZonMw) Project104003005, with additional support of Netherlands Organisation for Scientific Research (NWO), project NWO-EWVIDI 639.022.010 and project NWO-TTW Perspectief Programme P15-26. This work was partly carried out on the Dutch national e-infrastructure with the support of SURFCooperative. 

\end{document}